\documentclass{easychair} 
\usepackage{hyperref}
\usepackage{url}
\usepackage{graphicx}
\usepackage{natbib}
\usepackage[T1]{fontenc}
\usepackage[utf8]{inputenc}
\usepackage{nicefrac}
\usepackage{amssymb}
\usepackage{amsmath}
\usepackage{amsthm}
\usepackage{appendix}
\usepackage{graphicx}
\usepackage{subcaption}
\usepackage{multirow}

\DeclareMathOperator{\relu}{relu}
\DeclareMathOperator{\concat}{concat}

\title{Deep Network Guided Proof Search}

\author{Sarah Loos\inst{1} \and Geoffrey Irving\inst{1} \and Christian Szegedy\inst{1} \and Cezary Kaliszyk\inst{2}
}

\institute{
  Google Research \\
  \email{\{smloos|geoffreyi|szegedy\}@google.com}
\and
  University of Innsbruck, Austria\\
  \email{cezary.kaliszyk@uibk.ac.at}
}

\titlerunning{Deep Network Guided Proof Search}

\authorrunning{S.~Loos, G.~Irving, C.~Szegedy, and C.~Kaliszyk}

\begin{document}

\maketitle

\begin{abstract}
\noindent Deep learning techniques lie at the heart of several
significant AI advances in recent years including
object recognition and detection, image captioning,
machine translation, speech recognition and synthesis, and playing the
game of Go.

Automated first-order theorem provers can aid in the formalization and
verification of mathematical theorems and
play a crucial role in program analysis, theory reasoning, security,
interpolation, and system verification. 

Here we suggest
deep learning based guidance in the proof search of the theorem prover E.
We train and compare several deep neural network models on the traces of
existing ATP proofs of Mizar statements and use them to select processed clauses
during proof search.  
We give experimental evidence that with a hybrid,
two-phase approach, deep learning based guidance
can significantly reduce the average number of proof search
steps while increasing the number of theorems proved.

Using a few proof guidance strategies that leverage deep neural networks,
we have 
found first-order proofs of 7.36\% of the first-order logic translations of the
Mizar Mathematical
Library theorems that did not previously have ATP generated proofs. This
increases the ratio of statements in the corpus with ATP generated proofs from
56\% to 59\%.

\end{abstract}

\section{Introduction}

\subsection{Motivation}
In the past twenty years, various large corpora of computer-understandable
reasoning knowledge have been developed~\citep{HarrisonUW14}. Apart from
axioms, definitions, and conjectures, such corpora include proofs derived in
the selected logical foundation with sufficient
detail to be machine-checkable. This is either given in the form of
premises-conclusion pairs~\citep{Sutcliffe09a} or as procedures
and intermediate steps~\citep{Wenzel99}. The development of many of these formal
proofs required dozens of person-years, their sizes are measured in
tens of thousands of human-named theorems
and the complete
proofs contain billions of low-level inference steps.

These formal proof libraries are also interesting for AI-based methods,
with tasks such as concept matching, theory exploration, and structure
formation~\citep{AutexierH15}.
Furthermore, the AI methods can be augmented by automated reasoning: progress
in the development of efficient first-order \emph{automated theorem provers} (ATPs)~\citep{Vampire}
allows applying them not only as tools that redo the formal proofs, but also to
find the missing steps~\citep{Urban06}. Together with 
 proof translations
from the richer logics of the interactive systems to the simpler logics of the ATPs
this becomes a commonly used tool in certain interactive provers~\citep{BlanchetteKPU16}.
Many significant proof developments covering both mathematics and computer science
have been created using such technologies. Examples include the formal proof of the
Kepler conjecture~\citep{HalesABDHHKMMNNNOPRSTTTUVZ15}, or the proof of correctness
of the seL4 operating system kernel~\citep{KleinAEHCDEEKNSTW10}.

Despite the completeness of the employed proof calculi, modern ATPs perform
poorly in the presence of large fact libraries. For this reason AI-based heuristic
and learning techniques are used to preselect lemmas externally~\citep{KuhlweinLTUH12}.
Even with external selection~\citet{AlamaKU12} show that ATPs are only able to
find proofs of up to twenty human steps among the proofs in the Mizar Mathematical
Library~\citep{GrabowskiKN15},
whereas the library contains a number of proofs with hundreds
of steps.  In contrast to external lemma selection, guiding an ATP internally has much
better potential as the complete proof state is known.
For the tableaux calculus the \emph{Machine Learning Connection Prover}
(MaLeCoP)~\citep{UrbanVS11} shows that using machine learning to choose the next
step can reduce the number of inferences in the proofs on average by a factor of 20.
As the most competitive
ATPs today are not based on tableau, but instead rely on the superposition calculus,
in this paper we investigate guiding the state-of-the-art automated prover
E~\citep{Schulz13}\footnote{Version 1.9.1pre014} using deep neural networks.

Deep convolutional neural networks~\citep{lecun1998gradient} lie at the heart
of several recent AI breakthroughs in the past few years.
Deep convolutional networks have been instrumental in speech
recognition~\citep{hinton2012deep} and natural language
processing~\citep{kim2014convolutional}.
Object recognition has reached human level performance on large
benchmarks ~\citep{krizhevsky2012imagenet,szegedy2015going,he2015deep}
due to the inroads of deep convolutional neural networks. DeepMind's
AlphaGo~\citep{silver2016mastering} demonstrated superhuman performance
in playing the game of Go by utilizing deep convolution neural
networks that evaluate board positions. Deep hierarchical convolutional networks
have been responsible for vast improvements in speech and sound
generation~\citep{van2016wavenet}.

The wide applicability of deep neural architectures suggest their potential
usefulness for guiding the combinatorial search for proofs of mathematical
theorems as well.

\subsection{Contributions}
For the first time, we evaluate deep network models as a proof guidance method
inside an automated theorem prover. We describe the technicalities that need
to be addressed for a successful integration of relatively slow neural network
models in a system that relies on very fast 
exploration.

Experimental results are given on a large corpus of mathematical statements
translated from the Mizar Mathematical Library~\citep{mizar-in-a-nutshell}
to first-order logic~\citep{Urban06} which covers significant parts of basic
mathematics ranging from discrete mathematics and mathematical logic to
calculus and algebra.

Applying deep network models inside the proof search process is challenging
due to the relatively expensive nature of neural network inference.
During the time one clause is evaluated by a neural network (on CPU),
several hundreds or thousands of superposition steps may be performed by the
prover. This means that proof guidance via neural networks has to
provide very good quality suggestions to have a positive effect on the proof
process. In this paper we describe two important technical details that were
critical for getting improved results with neural network guidance.

First, interleaving network guidance and hard-coded heuristics
is still helpful for improving the performance over using the neural network
alone to pick the next processed clause.

Second, we use a two-phase
approach where a network-guided phase is followed by a
phase using only fast heuristics.  Without this second phase, the network-guided
approach often processes only 1\% of the clauses of a search with only
fast heuristics.  While the network-guided phase is slow, its high quality sets
up the rest of the search process at a better starting point.
In essence, if we spend half of the time guiding
the search and spend the rest finishing from that starting point,
we get significant improvements over lower-quality clause selection
that has a higher chance of losing its way in the search process and
omitting the selection of crucial clauses that are critical for finishing
the proof.

\section{Related Work}
\citet{suttner1990automatic} proposed a multilayer-perceptron based proof
search guiding algorithm on top of hand engineered features.
\citet{DenzingerFGS99} surveyed learning and knowledge
base based approaches to guide automated theorem provers.
\citet{Sch00} proposed a $k$-NN based learning methodology with hand-crafted
features on a normalized form of the clauses for the selecting the next clause
to be processed. Recently, for higher-order logic, na\"{\i}ve Bayes and particle swarm
based approaches were proposed for the internal guidance of the given-clause
algorithm in Satallax~\citep{faerber16mllax}. FEMaLeCoP uses na\"{\i}ve Bayes to
guide its tableaux proof search~\citep{KaliszykU15}. For the theorem prover {E},
a hand-engineered similarity function with a few learned weights was proposed
for better selection of processed clauses. All these approaches rely
on carefully engineered features, whereas our deep learning method
learns end-to-end from the textual representation or the syntax tree of the
proof clauses.

Recently, there have been several attempts to apply deep learning to theorem
proving and reasoning in general. Convolutional neural networks were used
successfully for premise selection in Mizar \citep{alemi2016deepmath},
and we use similar models for proof guidance here.
\citet{whalen2016holophrasm} proposed a complete neural network based
theorem prover architecture leveraging GRU~\citep{chung2015gated} networks to guide the
proof search of a tableau style proof process for the MetaMath
system~\citep{metamath}. A fully differential theorem prover
was tested on a few toy problems by~\citet{rocktaschel2016ntp}.
Entailment analysis via deep learning was proposed
by~\citet{rocktaschel2015reasoning}.

\section{Theorem Proving Preliminaries}

\subsection{The First-Order Logic Prover E}
\label{traces}
We target the saturation-based first-order logic theorem prover
E~\citep{Schulz13}.  E preprocesses all inputs into \emph{clausal normal form}
(CNF) and conducts the proof search in this form.  A first-order formula is in
CNF if it is a conjunction of clauses where each clause is a disjunction of
literals, where each literal is a (possibly negated) application of an $k$-ary
predicate symbol to $k$ terms, and where the terms are recursively either variables
or functions applied to terms.

E's proof search is parametrized by variety of arguments, the most important of
which is the \emph{clause selection heuristic} which we focus on in this work.
In saturation-based theorem proving two sets of clauses are manipulated:
a set of \emph{unprocessed clauses} initialized to the clausal form
of the problem and an initially empty set of \emph{processed clauses}. At each
step the heuristic chooses one of the unprocessed clauses, generates
all its consequences (adding them to the set of unprocessed clauses), then
moves the clause into the processed clauses set.  The order in which the
heuristic processes the clauses greatly impacts how long it will take
before a proof is found.  E's \texttt{auto} mode inspects a problem and selects
a clause search heuristic and other parameters likely to perform well.

The most successful heuristics in E are hybrid heuristics that select
the next clause by targeting different criteria in a round-robin fashion.  For
example, a simplistic hybrid heuristic might alternate between selecting
clauses in a FIFO ordering and selecting the shortest clause from the set of
unprocessed clauses.  These two heuristics in isolation may miss selecting
important clauses, but improve significantly when used in combination.  
The E prover implements hybrid heuristics by maintaining a separate ranking of all
unprocessed clauses for each of the selection criteria.  It then processes the
top clause from each ranking, allowing for an arbitrary
interleaving of selection criteria.

In this paper, we demonstrate that, given fixed time and
memory limits, we can prove more statements when we add
machine learning based classification to the clause
selection heuristic.

\subsection{Mizar First-Order Problems}
\label{mizar-fol}

The \emph{Mizar Mathematical Library} (MML)~\citep{GrabowskiKN15} is a library of
formal mathematics developed on top of the Mizar system~\citep{BancerekBGKMNPU15}.
The MML is today one of the
largest libraries of formalized proofs, covering most basic domains of mathematics and
some computer science proofs.
The foundations of Mizar have been translated to first-order logic by~\citet{Urban06}
making MML the first target for experiments combining machine learning with
automated reasoning.
The most extensive evaluation of AI-ATP methods on the library has been performed
by~\citet{KaliszykU13b} on MML version 4.181.1147. The set of 57,882 Mizar
theorems translated to first-order logic has been chronologically ordered
together with about 90,000 other formulas (mostly to express Mizar's type system).

The first-order problems used to guide E prover in this paper
are all distinct proofs found in~\citet{KaliszykU13b}. This includes both
proofs based on the human dependencies that the ATPs can redo and proofs
found by running the ATPs on the predicted dependencies. 
Of the 57,882 Mizar theorems in FOL, 32,521 (about 56\%) of them have at least one ATP proof
with a total of 91,877 distinct ATP proofs.

We use these 91,877 proofs to train and evaluate our neural networks.  
We randomly assign them by conjecture
into a training and a validation set using a $90\%$-$10\%$ split. 
By splitting the proofs by conjecture, we avoid contamination of our evaluation set.
If one conjecture has multiple proofs, then all of those proofs are 
assigned to the training set, or all are assigned to the evaluation set.  
The training set contains 82,718 unique proofs of 29,325 conjectures. 
The remaining 3,196 conjectures and their 9,159 proofs are assigned to the evaluation set.
When we evaluate clause selection heuristics over this evaluation set, we refer to them
as the {\it easy statements}, since they have previously been proved by some ATP method.
We call the 25,361 conjectures for which no ATP proof has been found the {\it hard statements}. 
These theorems taken together comprise the 57,882 Mizar theorems in FOL.

\section{Deep Networks}
\label{sec:deepnets}

Here we describe the data collection, network architectures, and training
methodology of our deep learning models to be incorporated into the
proof search procedure of the automated theorem prover E by \citet{Schulz13}.

\begin{figure}[t]
  \begin{subfigure}[t]{0.5\textwidth}
    \centering
    \includegraphics[width=0.8\textwidth]{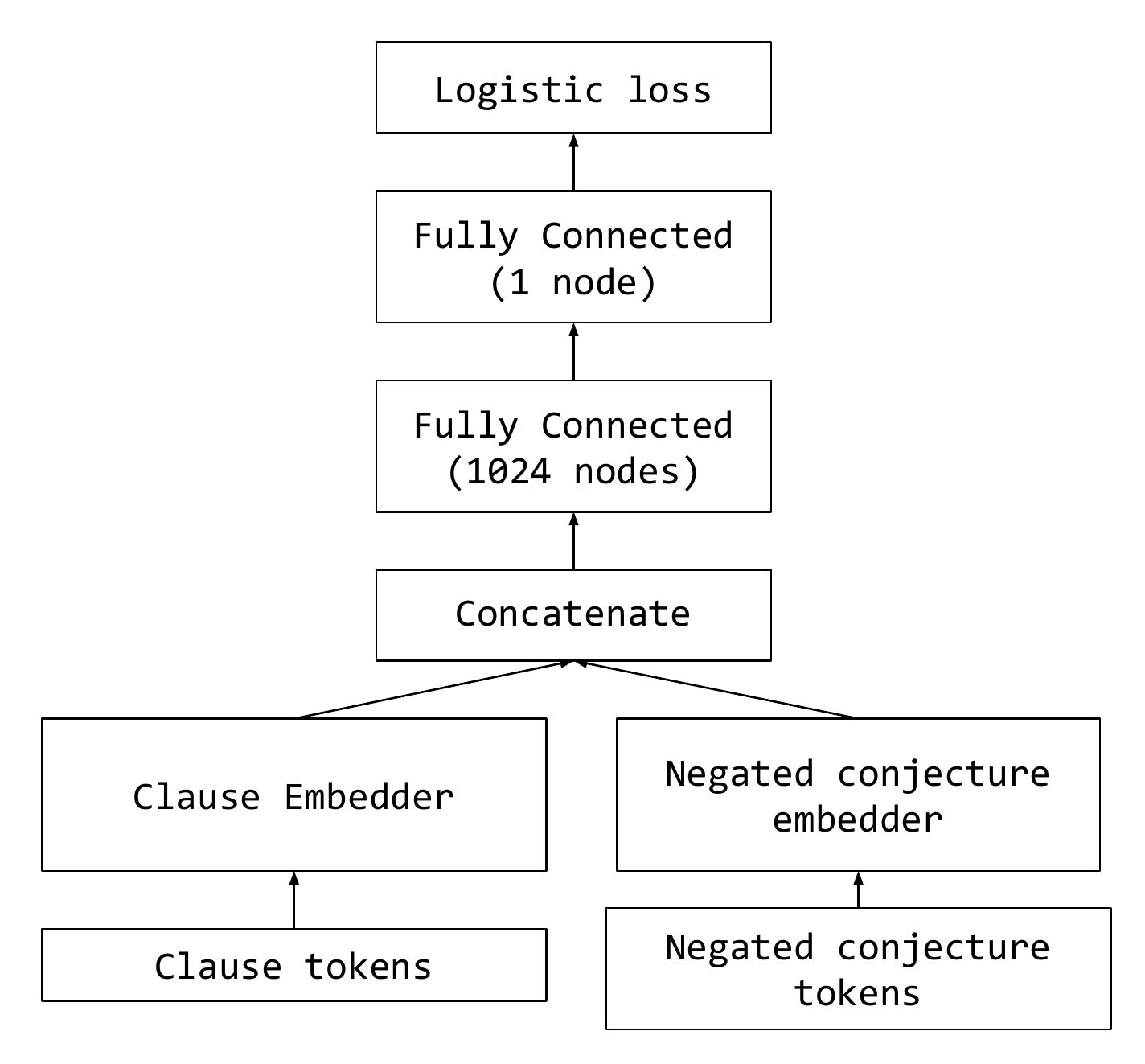}
  \end{subfigure}
  \begin{subfigure}[t]{0.5\textwidth}
    \centering
    \includegraphics[width=0.8\textwidth]{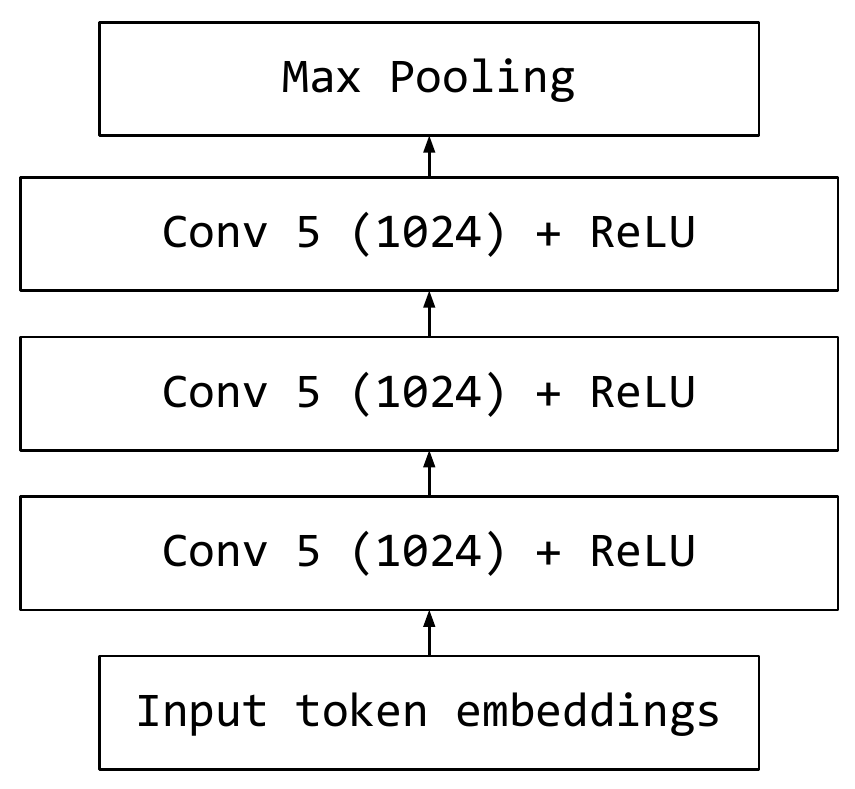}
  \end{subfigure}
\caption{Left: Overall network with clause and negated conjecture inputs.
Right: Convolutional embedding model.  Other experiments use WaveNet or recursive
embedding networks.}
    \label{convnet}
\end{figure}

We have reused several architectural and training choices that
proved successful in the earlier related premise selection work by
\citet{alemi2016deepmath}.  Our models have two inputs: a negated conjecture
to be proved and an unprocessed clause, both in CNF.  Each is reduced to a
fixed size vector using an embedding network, and a combiner network combines
the two embedding vectors into a score for use as a clause selection heuristic.
Ideally, the score would depend on the current set of processed clauses, but
we have restricted the inputs to negated conjecture and clause for simplicity
and speed.  Thus, the overall architecture is
\begin{align*}
v_\text{c} &= f_\text{emb}(\text{clause}, w_\text{c}) \\
v_\text{nc} &= f_\text{emb}(\text{negated-conjecture}, w_\text{nc}) \\
p(\text{useful} | \text{c}, \text{nc}) &=
    \sigma(g_\text{comb}(v_\text{c}, v_\text{nc}, w_\text{comb}))
\end{align*}
where $f_\text{emb}$ and $g_\text{comb}$ are the embedding and combiner
networks, $w_\text{c}$, $w_\text{nc}$, and $w_\text{comb}$ are the weights to
be learned, and $\sigma(z) = 1/(1 + e^{-z})$ is sigmoid (Figure~\ref{convnet},
left).  Note that we use the same architecture to embed both negated
conjecture and clause, but with different learned weights.

At training time, the output probabilities are trained using logistic loss
towards $p = 1$ if the clause was used in the proof, and towards $p = 0$
otherwise.  At proof time, the unprocessed clause with maximum probability is
chosen to be processed.

For the embedding network $f_\text{emb}$, we have tried three architectures:
\begin{enumerate}
  \item A simple convolutional network with a few convolutional layers
        followed by max-pooling.
  \item A WaveNet~\citep{van2016wavenet} style network that can effectively
        model long range dependencies.
  \item Recursive neural networks~\citep{goller1996learning} whose topology
        depends on the syntax tree of the formula to be evaluated.
\end{enumerate}

In sections~\ref{cnnsec},~\ref{wavenetsec}, and \ref{recursivesec},
we give a detailed description of the models evaluated and compare their
predictive performance after describing the training and evaluation data in
section~\ref{datasec}.  The end-to-end system was only evaluated using the
fastest models and those with very good prediction performance.
Later, we use the best models to generate new ATP proofs for the ``hard subset''
of the Mizar corpus where current ATP methods have failed.

\subsection{Data Collection}
\label{datasec}
As described in Section~\ref{mizar-fol}, we have split the 32,521 theorems
of Mizar that have at least one proof into a training and validation set using a 90\%-10\% split.
The proofs in our dataset originate from various first-order ATPs. Furthermore,
even for proof traces originating from E,
the configuration of the preprocessing (skolemization and clausification)
has a big effect on the the symbols that appear in the clauses and their shape.
Therefore, we have replicated the proofs using a single, consistent configuration of the
preprocessing procedure of E in order to avoid mismatch between our training
set and the clauses seen at test time when we guide the search with the trained
model.  Specifically, we train on proofs generated using the {\tt Auto208} configuration
detailed in the appendix.

\subsection{Architectures}

We consider three architectures for the embedding networks $f_\text{emb}$: simple
convolutional models, WaveNet~\citep{van2016wavenet} style models, and
recursive networks~\citep{goller1996learning}.  All experiments use the same
combiner network $g_\text{comb}$ with one 1024-unit hidden layer:
$$g_\text{comb}(v_\text{c}, v_\text{nc}) =
    W_2\relu(W_1 \concat(v_\text{c}, v_\text{nc}) + b_1) + b_2$$
where $W_1 \in \mathbb{R}^{1024 \times 2\dim v}$, $W_2 \in \mathbb{R}^{1 \times 1024}$,
$b_1 \in \mathbb{R}^{1024}$, and $b_2 \in \mathbb{R}$.

At data collection time, we collect all symbols (constants, function
names, variable names, logical operations, and parentheses) into a vocabulary
and the input layer converts each into an embedding vector of dimension
1024 using table lookup. The embeddings are initialized randomly and
learned during the training process.  We shared the same embedding vectors
between clause and negated conjecture embedding.  The convolutional and WaveNet
models receive clause and negated conjecture as sequences of these embeddings;
the recursive networks use embeddings only for names at leaves of the CNF trees.
Unlike \citet{alemi2016deepmath}, we avoid character-level embeddings
since proof search can involve very large clauses with large computational cost.

All models were trained using TensorFlow~\citep{tensorflow2015-whitepaper} and
the Adam~\citep{kingma2014adam} optimizer.

\subsection{Simple Convolutional Model}
\label{cnnsec}
Following the premise selection models of~\citet{alemi2016deepmath}, our
convolutional models (``CNNs'' for ``convolutional neural networks'') have
relatively shallow depth as they give good results on that related task.  They
consist of a stack of three one-dimensional convolutional layers, each layer
with patch size 5 (the number of inputs each output is connected to), stride 1,
and a rectified linear activation. We have tried several models with varying
feature dimensions.

\subsection{WaveNet Model}
\label{wavenetsec}
\begin{figure}[t]
  \centering
  \includegraphics[width=0.8\textwidth]{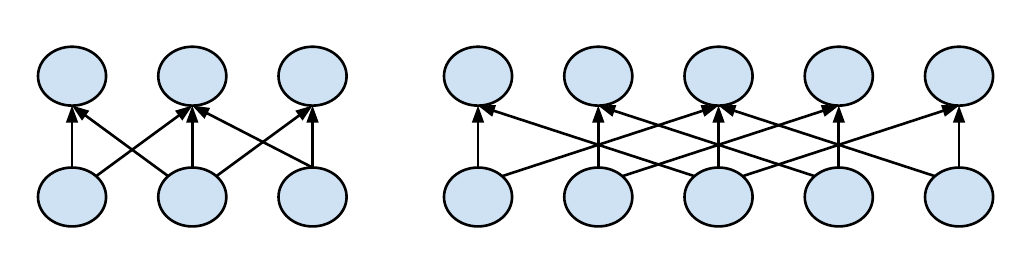}
  \caption{The difference between dilated and non-dilated convolution. On the
           left hand side is the schematics of a minimalist convolutional
           network with patch size $s=3$, stride $1$, whereas on the right is
           a dilated convolution with $s=3$, stride $1$, and dilation factor $d=2$.}
\label{fig:dilatedconv}
\end{figure}

WaveNet~\citep{van2016wavenet} is a special type of hierarchical convolutional
network that employs dilated convolutions and residual connections.  The
dilated convolutions allow long range feature dependencies with only moderate
overhead, as higher layers have geometrically increasing dilation factors (see
Figure~\ref{fig:dilatedconv}).  While~\citet{van2016wavenet} use causal dilated
convolutions, we use symmetric convolutions as our task is discrimination, not
generation.  Our WaveNet embedding network consists of 3 WaveNet blocks, where
each block consists of 7 convolutional layers dilated by $d = 1, 2, 4, \ldots,
64$.  We use the gated activation $\tanh(x) \sigma(x')$
of~\citet{van2016wavenet}, and residual connections for both layers and blocks
\citep{he2015deep}.  For regularization, some experiments use 20\% token-wise
dropout at the input layer and 20\% feature-wise dropout at the input to each
block~\citep{hinton2012improving}.  Overall,
\begin{align*}
f_{emb}(x) &= (B \circ B \circ B)(D_t(x,p)) \\
B(x) &= x + (L_{64} \circ \cdots \circ L_2 \circ L_1)(D_f(x,p)) \\
L_d(x) &= x + \tanh(C_d(x)) \sigma(C'_d(x)) \\
C_d(x)_i &= b + \sum_{j=1}^s w_j x_{i - d(j - \lceil s/2 \rceil)}
\end{align*}
where $D_t(x,p)$ sets each token embedding to zero with probability $p$,
$D_f(x,p)$ sets each individual feature to zero with probability $p$,
$C_d$ and $C'_d$ are convolutions with distinct weights and dilation
factor $d$, and $s = 3$ is the patch size of the convolution.  Here $x, B, L_d,
C_d$ are all sequences of vectors, or 2-D overall.  Our experiments use vectors
of dimension $256$ and $640$.

\subsection{Recursive Neural Network}
\label{recursivesec}
Our third model class is recursive neural networks \citep{socher2011parsing}
that are constructed by the parse tree of the first-order logic formulas.
The neural network topology reflects the parse tree and is wired on
the fly using {\tt gather} operations in TensorFlow.

To simplify the tree, all function applications are curried, giving
the following types of layers:
\begin{itemize}
  \item {\tt apply} nodes that apply a function with exactly two children.
  \item {\tt or} nodes that compute the embedding for the disjunction of exactly two children.
  \item {\tt and} nodes that compute the embedding for the conjunction of exactly two children.
    This is used only for embedding the negated conjecture, since the proof clauses do not
    contain conjunctions.
  \item {\tt not} nodes that compute the embedding for the negation of a single child node.
\end{itemize}

The weights of each type of layer is shared across the same tree. This means
that the layer weights are learned jointly for the same formula, as they
tend to contain multiple instances of the node of the same type.

At the leaves of the tree, we have the constants that can represent
functions of various arity. These symbols have their associated embeddings
which are initialized randomly and learned together with the rest of
the network.

\begin{figure}[t]
  \begin{center}
     \includegraphics[height=2.2in]{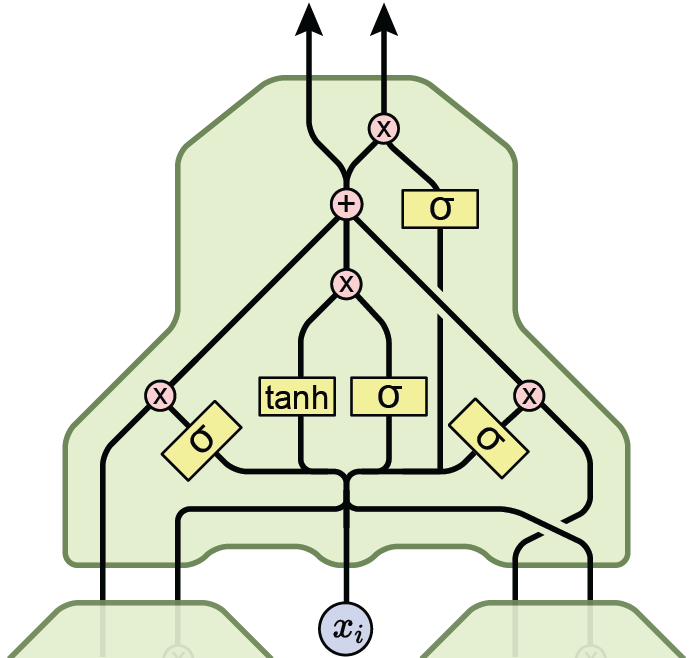}
     \qquad
     \includegraphics[height=2.2in]{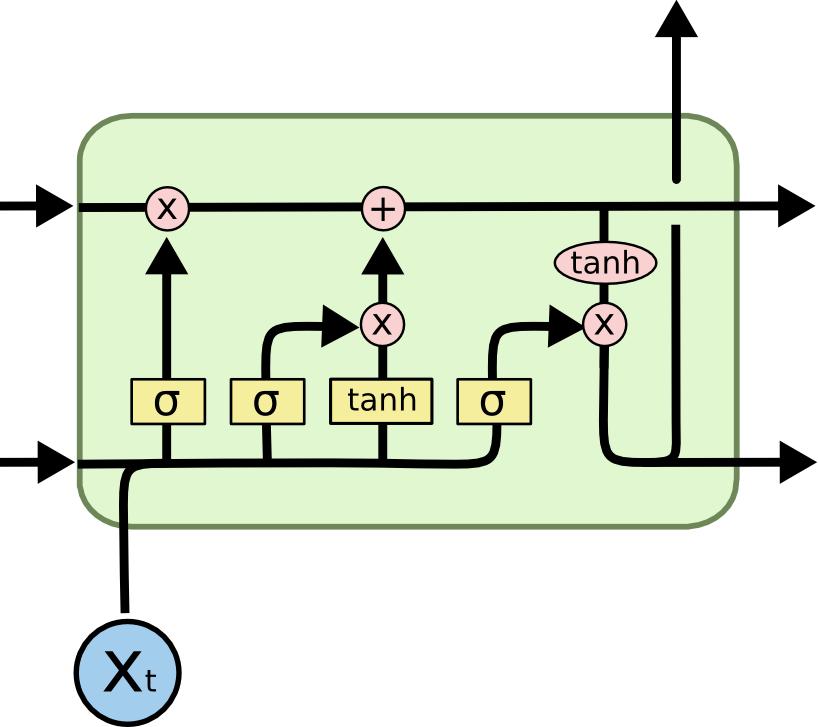}
  \end{center}
\caption{An tree LSTM block for a node with two children (left) compared with a
         conventional sequence LSTM (right).  For internal nodes, $x_i$ is
         empty in the first layer and equal to the output of the layer below for
         multilayer tree LSTMs.  Figures courtesy of \citet{olah2015lstm}.}
\label{fig:tree-lstm}
\end{figure}

Each of these internal nodes are represented by either a fully connected
layer with ReLU activation or tree LSTM network~\citep{tai2015improved}.
They aggregate the fixed size children and output a vector that has the length
of the pre-set output embedding size (or two vectors for tree LSTMs).
Our tree LSTMs use separate forget gains per input and include off-diagonal forget
gate terms (see Figure~\ref{fig:tree-lstm} and \citet{tai2015improved} for
details).

The network learns independent weight matrices for these three
(four for negated conjectures) transformations.
Although the same method is used for computing the embedding for the
proof-step clause and for the embedding, their weights
are not shared: we end up with seven sets of weights to learn in total.
We tested the common embedding sizes of 256 and 512. Greater embedding
size gave rise to more accurate models.

\section{Experimental Results}
\label{sec:experimental}

After training the deep neural networks as described in Section \ref{sec:deepnets}, 
we evaluate them using three measures, each more conclusive and
computationally expensive than the last.

The first metric, presented in Section \ref{sec:accuracy}, checks that a trained model 
can accurately predict whether a clause was used in the final proof.  This accuracy test 
is done on a holdout set of proof tasks from the training data.  

Next, in Section \ref{sec:easy}, we run E prover over the same holdout set of 9,159 FOL proof tasks
from~\citet{KaliszykU13b} using the trained networks to help guide the clause selection. 
These theorems all have at least one ATP proof
using classical search heuristics, and $96.6\%$ of these theorems can be
proved using one of E prover's built-in automated heuristics, so there is little room
for improvement on this dataset.  
However, it allows us to perform a sanity check that the neural nets aren't doing
more harm than good when added to the auto heuristic. 

The final test is the most interesting but computationally expensive: do these 
deep network guided selection heuristics allow us to prove Mizar statements 
that do not yet have any ATP proof? 
For this, we use a corpus of 25,361 theorems from Mizar for which no proof was found in \citet{KaliszykU13b}.
Because this is the most computationally expensive task, we only run it using
the networks that performed best on the previous tasks.
In Section \ref{sec:hard}, we present the models that, in aggregate, allow us to find proofs for $7.36\%$ 
of these ``{\it hard statements}''.

\subsection{Accuracy Evaluations}
\label{sec:accuracy}
\begin{table}
\begin{center}
                \begin{tabular}[H]{|r|c|c|}
                   \hline
           {\bf Model} & {\bf Embedding Size} & {\bf Accuracy on 50-50\% split} \\ \hline
           Tree-RNN-256$\times$2 & 256 & 77.5\% \\ \hline
           Tree-RNN-512$\times$1 & 256 & 78.1\% \\ \hline
           Tree-LSTM-256$\times$2 & 256 & 77.0\% \\ \hline
           Tree-LSTM-256$\times$3 & 256 & 77.0\% \\ \hline
           Tree-LSTM-512$\times$2 & 256 & 77.9\% \\ \hline
           CNN-1024$\times$3 & 256 & 80.3\% \\ \hline
           *CNN-1024$\times$3 & 256 & 78.7\% \\ \hline 
           CNN-1024$\times$3 & 512 & 79.7\% \\ \hline
           CNN-1024$\times$3 & 1024 & 79.8\% \\ \hline
           WaveNet-256$\times$3$\times$7 & 256 & 79.9\% \\ \hline
           *WaveNet-256$\times$3$\times$7 & 256 & 79.9\% \\ \hline 
	   WaveNet-1024$\times$3$\times$7 & 1024 & 81.0\% \\ \hline
	   WaveNet-640$\times$3$\times$7$(20\%)$ & 640 & \textbf{81.5}\% \\ \hline
           *WaveNet-640$\times$3$\times$7$(20\%)$ & 640 & 79.9\% \\ \hline 
        \end{tabular} 
\end{center}
  \caption{The accuracy of predicting whether a processed clause was ultimately required
           for the proof. The accuracy is measured on a 50-50\%
           split of positive and negative processed clause examples, with various
           recursive deep neural network models.
           Models with an asterisk (*) were trained on
           a data set which additionally included a sampling of unprocessed clauses as
           negative examples.  In order to facilitate a direct comparison with other models, the 
           same evaluation dataset was used, but this is slightly 
           biased against the examples denoted with (*). }
  \label{tab:results}
\end{table}

Table~\ref{tab:results} shows the accuracy on a 50-50\% split of used and unused
clauses of the holdout set of the conjectures.  Here CNN-$N$$\times$$L$ is a
convolutional net with $L$ layers of $N$ dimensions, and
Tree-\emph{Type}-$N$$\times$$L$ is a tree RNN or LSTM with $L$ layers
of $N$ dimensions at each node of the input tree.  
WaveNet-$N$$\times$$B$$\times$$L$ has $B$ blocks, with $L$ layers, 
of dimension $N$.  We include $(D\%)$ to indicate that a dropout of $D\%$ 
is used as a regularizer during training.

The WaveNet 640 with dropout has the best accuracy at 81.5\% accuracy, 
but many of the CNN and WaveNet models noticeably outperform the others. 
Note that the time required for
evaluating the models for a set of examples varies widely. Given the fact
that we limit the overall running time instead of the number of evaluations
performed, higher quality but slower models might perform better in the
system than slightly worse but much faster models. However with specialized
hardware for neural network evaluation, we can expect that the prediction
quality of the models will gain in importance compared to their running time.

\subsection{Experiments on Statements with Existing ATP Proofs}
\label{sec:easy}

The clause selection heuristic is one of the most important parts of E prover,
as it is the primary driver of the proof search.
With a good heuristic, a proof may be found in a relatively small number of
search steps. In this section, we use the models from Section
\ref{sec:accuracy} with the highest accuracy as the clause
selection heuristic inside of E.\footnote{Due to its huge memory footprint, 
we were unable to experiment with WaveNet-1024 in E prover.}
These models, which are trained to select
clauses that are most likely to contribute to a proof, now assign a score
to each unprocessed clause, namely the trained value of $p(\text{useful}| \text{c,nc})$.
In other words, the probability that a clause c, is used in the final proof, given the set of negated conjectures nc. 
E prover uses this score to rank the set of unprocessed clauses, then at each step 
processes the clause that has the best ranking.\footnote{E prover uses a lowest-is-best ordering, so in implementation we use $-p(\text{useful}| c,nc).$}
Since many new unprocessed clauses are generated when each clause is processed,
this selection order is crucial.  A good clause selection heuristic can be the
difference between finding a proof after a small number of search steps, and
finding a proof only after years of computation time.

While our ultimate goal is to develop a heuristic that is powerful enough to prove
the challenging statements that don't yet have an ATP proof, experimenting on
this dataset is computationally very expensive.  Instead, we kept a holdout
set of 9,159 statements which we did not use as training data.  These
statements already have existing ATP proofs and tend to be
computationally less intensive, so we can use this holdout set to run more experiments
and directly compare with proofs generated by an existing Auto strategy, 
which we selected as the best-performing strategy on this dataset as discussed in 
the appendix.

In this section, we find that deep neural nets, which are computationally very expensive
even when used for inference, are most effective when used in combination with
existing heuristics.
We also use the holdout set to investigate which of the model architectures presented in
Section~\ref{sec:deepnets} are most effective.

\subsubsection{Guidance Design}
\label{sec:design}

\begin{figure}
  \begin{center}
     \includegraphics[width=0.45\textwidth]{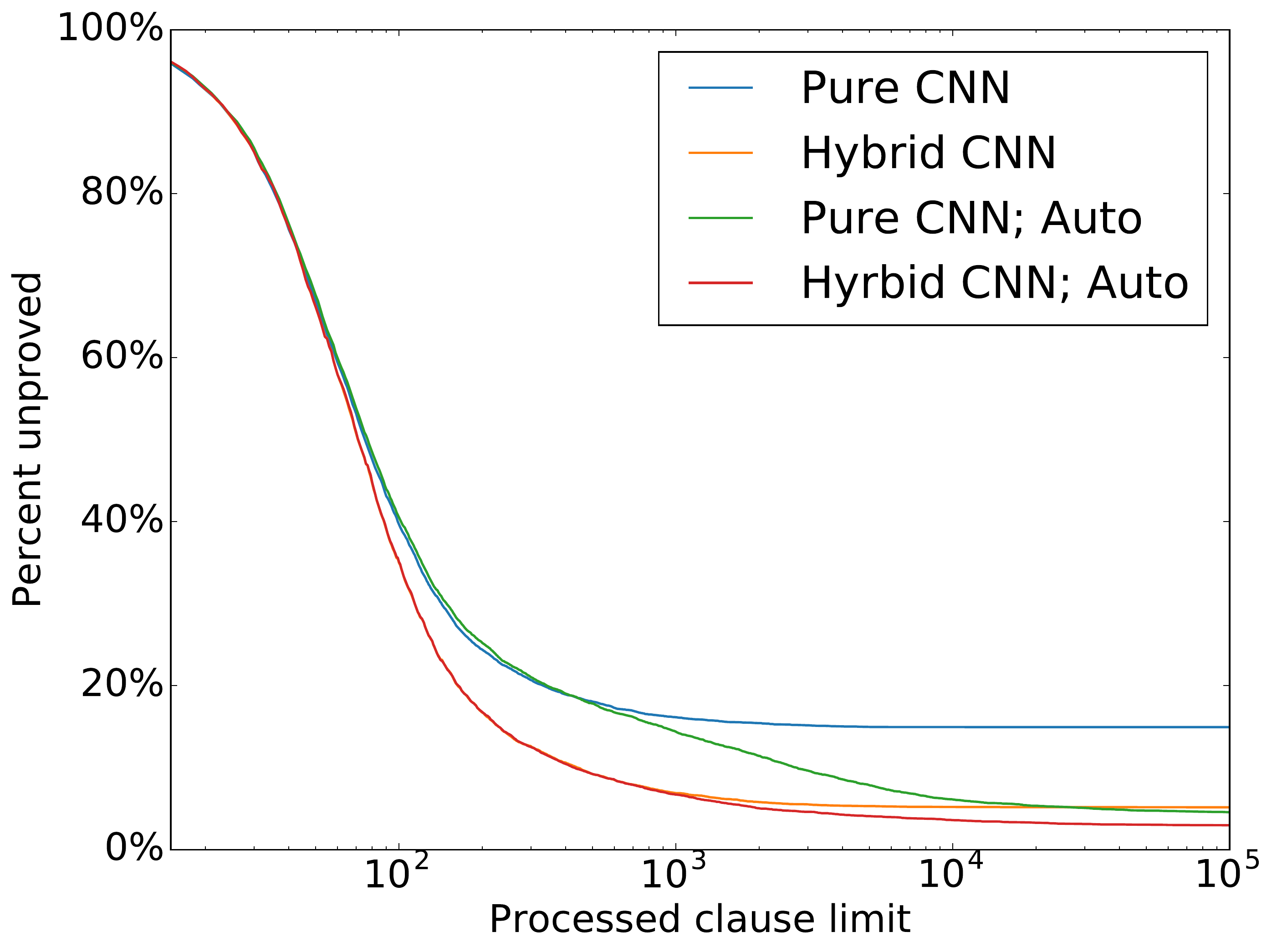}
     \qquad
     \includegraphics[width=0.45\textwidth]{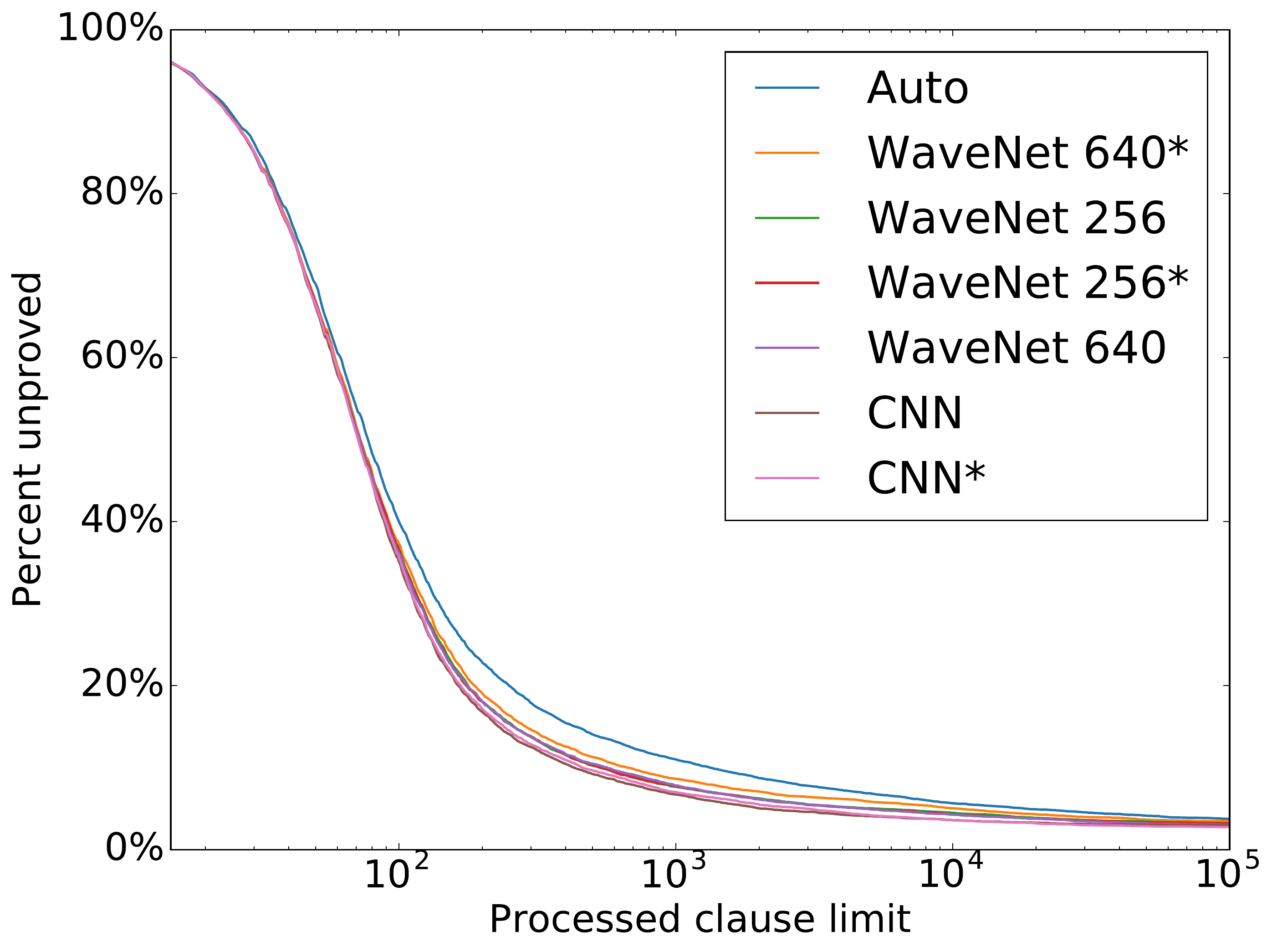}
  \end{center}
\caption{The percentage of unsuccessful proofs at various
         processed clause limits using different selection heuristics.
         In the left figure, we use the same network (CNN-1024x3, with 256 embedding) throughout,
         but show the effect of various interactions with the Auto heuristic.
         On the right we use hybrid, two-phase guidance, but show the
         effect of different neural networks.  More detailed values are shown
         in Table \ref{tab:percent_proved}.}
\label{fig:guidance_design}
\end{figure}

In our experiments, we found that our computation heavy neural networks did much better when they 
were used in concert with existing, fast heuristics.  In this section, we present different approaches
to using neural networks to guide existing heuristics.  
\begin{enumerate}
\item The Auto heuristic is standard in E (see appendix).
	This approach is already a hybrid of several weighting functions and tuned parameters, but does 
	not include any inference from a deep neural network. 
\item A pure neural network heuristic scores the clauses using only the trained neural network.  
	This approach is slow, as it must evaluate all clauses with the expensive neural network. 
	It also can not take advantage of ranking based on different heuristics, as the Auto function does, 
	and we observe that it does not perform well on its own.	
\item A hybrid approach alternates between deep network based
        guidance and the fast Auto heuristic. This
        augments the standard clause selection methods with additional
        neural network suggested clauses. While this process still can do relatively
        few proof search steps due to the slowness of the neural network
        evaluations, which must still be done on all unprocessed clauses 
        to obtain a rank ordering.
\item A switched approach, which uses deep network guidance in the first phase of the computation (either pure or hybrid).
	As time resources begin to run out, we switch to Auto, a traditional complete 
	search phase, which can process a lot of clauses without the overhead of the deep network guidance.
\end{enumerate} 
Figure \ref{fig:guidance_design} left shows a direct comparison of these approaches to proof
guidance using a simple CNN in all cases.  
Not surprisingly, the pure CNN does not perform well on its own; however, 
when we allow E prover to alternate between the CNN
heuristic and the Auto heuristic, the hybrid
outperforms both the Auto and CNN heuristics for lower limits on the processed
clauses.
Still, because the hybrid approach has all the overhead of the deep network evaluation, 
we see this hybrid method bottoms out after around 7,500 processed clauses, due 
to a lack of resources.

The best of the guidance approaches we explored was the switched approach,
which uses the Auto heuristic to avoid running out of resources at a relatively 
small number of processed clauses.
The intuition behind this is that we expect the standard
heuristics to be too weak to avoid combinatorial explosion while neural network
guidance is still too slow on current hardware. If we run a guided proof
search alone for awhile, it could end up selecting all the essential clauses
for the proof, but it might fail to close to proof due to its slowness.
Our ``switched'' approach gives a cure to this shortfall of the network guided
approach.

Throughout the remainder of this paper, all the proof guidance methods employ
the two-phase ``switched'' approach that
first runs a hybrid network-guided phases for 20 minutes, followed by
a standard heuristics-based phase for 10 minutes.

\subsubsection{Comparison of Model Performance on Easy Statements}

\begin{table}
\begin{center}
\begin{tabular}[H]{|r|c|c|c|c|c|c|}
	\hline
      {\bf Model} &  {\bf Accuracy} & {\bf PC$\boldsymbol\leq$1,000} & {\bf PC$\boldsymbol\leq$10,000} & {\bf PC$\boldsymbol\leq$100,000} & {\bf PC$\boldsymbol<$$\boldsymbol\infty$} \\ \hline
      Auto             &  N/A & 89.0\% & 94.3\% & 96.2\% & 96.6\% \\ \hline \hline
      *WaveNet 640     & 79.9\% & 91.4\% & 95.0\% & 96.5\% & 96.6\% \\ \hline
      WaveNet 256      & 79.9\% & 92.3\% & 95.5\% & 96.8\% & 96.8\% \\ \hline
      *WaveNet 256     & 79.9\% & 92.2\% & 95.7\% & 96.8\% & 96.8\% \\ \hline
      CNN              & 80.3\% & {\bf 93.3}\% & {\bf 96.4}\% & 97.0\% & 97.1\% \\ \hline
      WaveNet 640      & \textbf{81.5}\%  & 92.2\% & 95.7\% & 97.0\% & 97.2\% \\ \hline
      *CNN             &  78.7\% & 93.0\% & {\bf 96.4}\% & {\bf 97.2}\% & {\bf 97.3}\% \\ \hline
      \end{tabular}
\end{center}
\caption{The percent of theorems proved in the ``{\it easy statements}'' with various model architectures.
              The training accuracy is duplicated from Table \ref{tab:results} for convenience.
              The values in the PC$\leq$$N$ columns indicate the percent of statements proved requiring fewer than $N$ 
              processed clauses (PC).  The far right column (PC$<$$\infty$) is the percent of statements proved within
              30 minutes, with a memory limit of 16G, and no limit on processed clauses.}
\label{tab:percent_proved}
\end{table}

In this section we present the performance of four WaveNet and two CNN models 
used to guide clause selection  in E prover. 
These are the models with the highest accuracy from Table \ref{tab:results}, which are 
still small enough to fit in memory.
We also include all three models trained on a data set that includes unprocessed clauses 
as negative examples, these models are indicated with a (*).
All experiments use the E theorem prover,
version 1.9.1pre014~\citep{Schulz13}, with some modifications to enable
integration with the trained deep neural networks as clause selection heuristics.
 
In Figure \ref{fig:guidance_design} right and Table \ref{tab:percent_proved},
we show the performance of each of these models with various processed clause limits. 
All of the heuristics that benefit from deep neural network guidance outperform the 
Auto baseline, regardless of the number of clauses processed before a proof is found.  
Both CNNs (embedding size 256) do very well, but we notice that the CNN* trained with unprocessed
clauses as negative examples begins to do better with higher processed clause limits, 
which may better represent the clauses that are evaluated after several thousand 
clauses are already processed.
Somewhat surprisingly, the WaveNet models (embedding size 256 and 640), 
which were more accurate at predicting whether a clause
would be used in the final proof, did not perform as well at the proof guidance task.
This is likely because the WaveNet models are much more resource heavy, and therefore 
could not evaluate as many clauses as the CNNs given the same resources.

While these results indicate that the that guidance from deep neural networks can provide some 
boost to traditional search heuristics, the Auto heuristic alone already performs very well on this
dataset, so the improvements are minor.
The real test is on the set of ``{\it hard statements},'' which did not previously have ATP
generated proofs. 


\subsection{Experiments on Hard Statements}
\label{sec:hard}
Here we describe our results on a corpus of 25,361 theorems from the Mizar corpus for
which no proof was found by any of the provers, strategies, and premise selection
methods considered by~\citet{KaliszykU13b}. We will call these
{\it ``hard statements''} in the rest of the paper.

Although all the hard statements have human proofs, this subset could neither
be proved by theorem prover E~\citep{Schulz13} nor by Vampire~\citep{Vampire}
with a timeout limit of 15 minutes and default settings
(auto and casc heuristics respectively) using the premises derived from the human
given proofs. Also note that the premises passed to the theorem prover form a
very rough over-approximation of the set of necessary premises. Often
around 10 to 20 dependencies are enough to prove the theorems, but the
number of dependencies in the backward envelope can be in the thousands.

The main bottleneck for network guided proof search is the evaluation time for
the ranking of clauses to be processed next. Given that we do not use special
hardware for this purpose, our proof search is completely dominated
by the deep network evaluation.

If the number of premises is very high in the beginning, there is a much
bigger hit on the deep network guided proof search due to the
larger number of clauses evaluations per step. This motivates the
use of additional premise selection as an important additional component.

Still, we first present a baseline without premise selection and show that
the relatively fast premise selection phase is critical for
good performance on the hard theorems with or without deep network
guidance. In the second subsection, we turn on premise selection and
compare various proof guidance models against the baseline auto heuristic.

\subsubsection{The Importance of Premise Selection}

Here we start with a comparison of four different approaches on the hard
statements. These include guided and unguided proof search with and without
premise selection.

For premise selection, we use the character level model
from the DeepMath paper~\citep{alemi2016deepmath}. The model is a very similar
model to our convolutional proof guidance model, but was trained on
premise-conclusion pairs from a training set that was randomly chosen
from the 56\% ATP-proved statements. One main difference is that our
proof guidance convolutional network is word level with an embedding
learned for each token, whereas the premise selection network
takes character sequences as input. This limits quality, as the DeepMath paper
suggests that using a word level model
with definitional embeddings gives the best results. We have
opted for the lower quality character level premise selection model for
simplicity.

After premise selection we run four proof attempts: first the premises
are ranked by the model scores and the top $32, 64, 128$, and $256$
different premises selected (as long as the original number of premises is not
exceeded). We run E prover using the selected set of premises as long as it
does not find a proof. We stop searching for a proof when a proof is found for
any subset of the premises.

The experimental results are given in Table~\ref{tab:selectionimportance}.
We give experimental evidence that the switched approach outperforms
the unguided proof search. We have tested two different premise selection
models to get a feel for the variability and complementarity of the results.
It turns out that different premise selection models even with the same
architecture can introduce a lot of variations that can lead to different
number of theorems proved, but can result in choices that complement each
other well and help the proof search process by increasing the diversity
of starting set of premises.

\begin{table}
\begin{center}
  \begin{tabular}[h]{| r | c | c | }
    \hline
    & without premise selection & with premise selection \\
    \hline\hline
    unguided & 145 & 458 \\
    \hline
    guided (hybrid) & 137 & 383 \\
    \hline
  \end{tabular}
\end{center}
\caption{Number of hard theorems proved with various combinations of
  premise selection and (unswitched) proof guidance. Note that even when our proof
  guidance is partial, it still produces worse results than the variant without
  deep network guidance. This is due to the slowness of deep network evaluation.
  The sole purpose of this table is to highlight the importance of premise
  selection for the hard statements. In other experiments, we concentrate on the
  two-phase ``switched'' approach that combines guided and unguided search in a
  sequential fashion and outperforms both unguided search and hybrid guidance
  without the switch.}
\label{tab:selectionimportance}
\end{table}

\subsubsection{Comparison of Model Performance on Hard Statements}

All the proof guidance methods employed the two-phase approach that
first runs a hybrid network-guided phases for 20 minutes, followed by
a standard heuristics based phase for 10 minutes.

Here we tried two different premise selection models ``DeepMath 1'' and
``DeepMath 2'' using the same character level convolutional architecture as
the best character level model described in ~\citep{alemi2016deepmath}.
This way, we can evaluate the stability of result with respect to the
premise selection strategy. One can see that although the two models are
trained the same way and have comparable quality, they result in significantly
different sets of theorems proved, but the number of theorems proved
using both models is very similar and the ``switch'' strategy using
the simpler CNN performed best for both premise selection and proof 
guidance by a significant margin.

The experimental results are given in Table~\ref{tab:hardresults}.
The CNN models use
a simple three layer convolutional networks to rank the unprocessed clauses
while the WaveNet models use the WaveNet architecture which is significantly
slower to evaluate, but produced better proxy metrics on the holdout set.
Note that the simple convolutional network based approach proves $86$\% more
theorems than the unguided approach ($4.34$\% versus $2.33$\% of the hard
statements). 
The overall number the statements proved by any means
in this paper (including the use of non-switched neural guidance) is 1,866
which is $7.36$\% of all the hard statements. 

\begin{table}
\begin{center}
  \begin{tabular}[h]{| r | c | c | c |}
    \hline
    {\bf Model} & {\bf DeepMath 1} & {\bf DeepMath 2} & {\bf Union of 1 and 2}\\
    \hline\hline
    Auto & 578 & 581 & 674 \\
    \hline
    *WaveNet 640 & 644 & 612 & 767 \\
    \hline
    *WaveNet 256 & 692 & 712 & 864\\
    \hline
    WaveNet 640 & 629 & 685 & 997 \\
    \hline
    *CNN & 905 & 812 & 1,057 \\
    \hline
    CNN & 839 & 935 & 1,101  \\
    \hline \hline 
    Total (unique)& 1,451  & 1,458 & 1,712 \\
    \hline
  \end{tabular}
\end{center}
\caption{Number of theorems proved out of the 25,361 hard theorems,
  proved with various combinations of premise selection (DeepMath 1 \& 2) and 
  clause selection guidance. The last column shows the union of  theorems proved with either
  premise selection step method in the given row.
  The size of the union of the all theorems proved by methods in this in this
  table is 1,712 (6.8\%).
  The number of theorems proved by the deep network guided methods is 1,665 (6.6\%).
}
\label{tab:hardresults}
\end{table}

\section{Conclusion}
We have demonstrated the feasibility of guiding first-order logic proof search
by augmenting the given clause selection method with ranking by deep neural networks.
With a properly engineered mixture of neural guidance and
hand-crafted search strategies, we get significant improvements in first-order
logic prover performance, especially for theorems that are harder and
require deeper search.

Due to the slowness of neural network evaluation, our method leads to
an increased number of successful proofs only if we utilize it in a
two-phase approach where a deep network guided phase is followed by a
faster combinatorial search with given clauses selected quickly
using the existing hand-crafted clause selection strategies.

Additionally, we established that both the prediction accuracy of the deep
network and its speed are important for increased proving power. For example,
our WaveNet model yields higher accuracy on the holdout set than a simple
convolutional network, but due to its much higher computational cost,
when used as a clause scorer in {E} prover, it still proves fewer theorems under
the same time constraints than the cheaper but lower quality convolutional network.

Currently our approach is computationally expensive, and we allow 30 minutes
per proof compared to 15 minutes for previous work.  However, this extra time
does not significantly help existing heuristics unless
combined with neural proof guidance (see appendix).
This suggests that the neural guidance represents a substantial improvement
in constraining the search space. Moreover, we expect that the inroads of
specialized hardware for deep learning will mitigate much of the
computational overhead. This would allow higher performing models to
play their strength with dramatic efficiency and quality gains in the future.

Our approach is just the first step of applying deep learning to guiding
combinatorial search processes and it is arguable
that working with a purely syntactic input format the derived set of features
is not strong enough to create a semantically relevant
representation of mathematical content.
Besides improving theorem proving, our approach has the exciting potential
to generate higher quality training data for systems that study
the behavior of formulas under a set of logical transformations.
This could enable learning representation of formulas in ways that consider
the semantics not just the syntactic properties of mathematical content and can
make decisions based on their behavior during proof search.

\bibliographystyle{iclr2017_conference}

\bibliography{clause_search}

\clearpage
\appendix
\appendixpage

\section*{Selecting Auto Baseline}

Here we compare several hybrid selection heuristics that are built into E prover.
In the paper, we use as a baseline the hybrid heuristic that performs best on our holdout set. 

First we run E prover on our holdout set with the {\tt \--\--auto} flag. 
When this flag is enabled, E prover chooses a (hybrid) selection heuristic dynamically based on features of the conjecture.
It also dynamically chooses a term ordering and a literal selection strategy \citep{emanual}. 

\begin{figure}[h!]
  \begin{center}
     \includegraphics[width=0.8\textwidth]{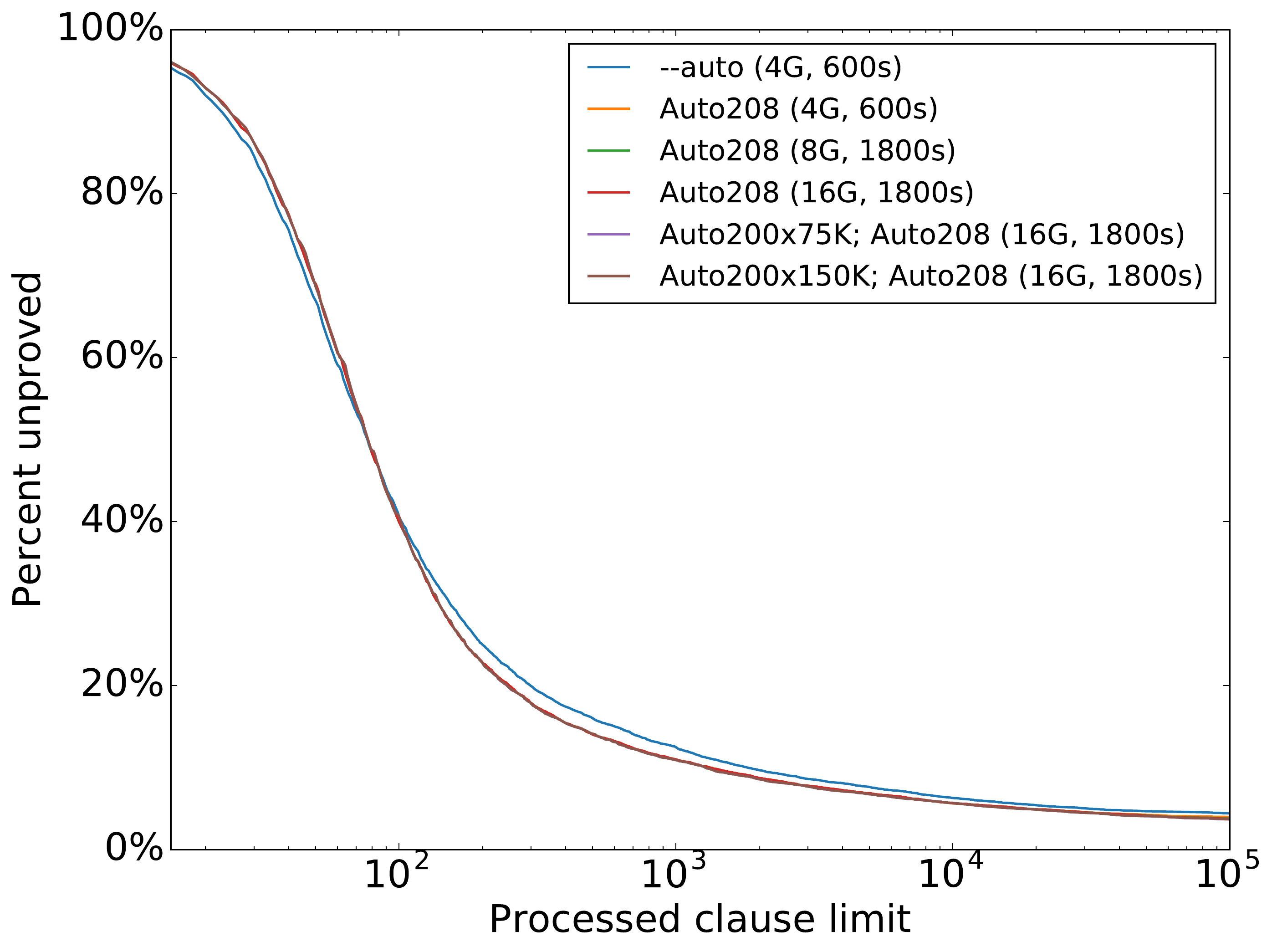}
  \end{center}
\caption{The percentage of unsuccessful proofs at various 
         processed clause limits using E prover selection heuristics. 
         Final values with no processed clause limit are presented in 
         Table~\ref{tab:auto_baseline}.}
\label{fig:baseline}
\end{figure}

We found that in $52.5\%$ of the proofs generated with the {\tt \--\--auto} flag, the {\tt Auto208} heuristic is used.   
When we use only the {\tt Auto208} heuristic on the full dataset with 4G of memory and a 600 second timeout, 
it proves more theorems than the {\tt \--\--auto} flag, as shown in Figure \ref{fig:baseline}.  

In Section \ref{sec:design}, we introduce a switched approach, which first runs a hybrid heuristic, 
and then as the theorem prover runs out of resources, switches to the {\tt Auto208} selection heuristic. 
To ensure that any benefit we gain is not due to the switching itself, we also experiment here with 
switching between two built-in heuristics. 
The {\tt \--\--auto} flag uses the {\tt Auto200} heuristic in $5.9\%$ of proofs on our holdout set, making it 
the second-most used heuristic. 
In Table \ref{tab:auto_baseline} we show that switching from the {\tt Auto200} to the {\tt Auto208} heuristic
has very little impact on the final result. 
The "{\tt Auto200}x150K; {\tt Auto208}" first runs {\tt Auto200} for up to 150K processed clauses, 
then runs {\tt Auto208} for the remainder (similarly for 75K). 
We select 150K processed clauses, as it is the median number of processed clauses for unsuccessful 
proofs using {\tt Auto208}.

\begin{table}[t]
\begin{center}
\begin{tabular}[H]{|r|c|c|c|c|c|}
      \hline
      {\bf Selection Heuristic} &  {\bf Memory} & {\bf Timeout} & {\bf Percent Proved} \\ \hline 
      {\tt \--\--auto} & 4G & 600s & 95.72\% \\ \hline
      Auto208 & 4G & 600s & 96.12\% \\ \hline
      Auto208 & 8G & 1800s & 96.52\% \\ \hline
      Auto200x150K; Auto208 & 16G & 1800s & 96.60\% \\ \hline
      Auto200x75K; Auto208 & 16G & 1800s &  96.62\% \\ \hline
      Auto208 &16G & 1800s &  96.64\% \\ \hline
\end{tabular}
\end{center}
\caption{Comparing auto baselines on the easy theorems.}
\label{tab:auto_baseline}
\end{table}

The full names and hybrid ordering of the {\tt Auto208} and {\tt Auto200} heuristics 
are included here, and with them we use term-ordering {\tt KBO6}.
Throughout this paper, we compare against {\tt Auto208} as our Auto baseline, 
since it was the best-performing heuristic in these experiments. 

\begin{verbatim}
"Auto208"
G_E___208_C18_F1_SE_CS_SP_PS_S0Y := 
1*ConjectureRelativeSymbolWeight(SimulateSOS,0.5,100,100,100,100,1.5,1.5,1),
4*ConjectureRelativeSymbolWeight(ConstPrio,0.1,100,100,100,100,1.5,1.5,1.5),
1*FIFOWeight(PreferProcessed),
1*ConjectureRelativeSymbolWeight(PreferNonGoals,0.5,100,100,100,100,1.5,1.5,1),
4*Refinedweight(SimulateSOS,3,2,2,1.5,2)
\end{verbatim} 

\begin{verbatim}
"Auto200"
G_E___200_C45_F1_AE_CS_SP_PI_S0Y :=
1*ConjectureRelativeSymbolWeight(SimulateSOS,0.5,100,100,100,100,1.5,1.5,1),
6*ConjectureRelativeSymbolWeight(ConstPrio,0.1,100,100,100,100,1.5,1.5,1.5),
2*FIFOWeight(PreferProcessed),
1*ConjectureRelativeSymbolWeight(PreferNonGoals,0.5,100,100,100,100,1.5,1.5,1),
8*Refinedweight(SimulateSOS,1,1,2,1.5,2)
\end{verbatim}

\end{document}